\documentclass[sigconf,pbalance=true]{acmart}

\usepackage{xurl} 
\usepackage{booktabs}
\usepackage{multirow}
\usepackage{graphicx}
\usepackage[table]{xcolor} 
\usepackage{arydshln}
\AtBeginDocument{%
  }

\copyrightyear{2026}
\acmYear{2026}
\setcopyright{cc}
\setcctype{by}
\acmConference[MM '26]{Proceedings of the 34th ACM International Conference on Multimedia}{November 10--14, 2026}{Rio de Janeiro, Brazil}
\acmBooktitle{Proceedings of the 34th ACM International Conference on Multimedia (MM '26), November 10--14, 2026, Rio de Janeiro, Brazil}
\acmDOI{10.1145/3767308.3836147}
\acmISBN{979-8-4007-2213-4/2026/11}

\begin{document}

\title{ADGNet: Asymmetric Dual-text Guided Network for Infrared Small Target Detection}


\author{Tongtong Wang}
\orcid{0009-0003-2421-1363}
\affiliation{%
  \institution{\normalsize Shandong University}
  \city{Jinan}
  \country{China}
}
\email{wangttong@mail.sdu.edu.cn}

\author{Mingzhu Xu}
\orcid{0000-0002-1492-0970}
\authornote{Corresponding author: Mingzhu Xu.}
\affiliation{
  \institution{\normalsize Shandong University}
  \city{Jinan}
  \country{China}
  }
\email{xumingzhu@sdu.edu.cn}

\author{Chenglong Yu}
\orcid{0009-0004-7729-1616}
\affiliation{
  \institution{\normalsize Shandong University}
  \city{Jinan}
  \country{China}
  }
\email{yucl@mail.sdu.edu.cn}

\author{Jing Wang}
\orcid{0009-0003-1107-2528}
\affiliation{%
  \institution{\normalsize Shandong University}
  \city{Jinan}
  \country{China}
}
\email{202415291@mail.sdu.edu.cn}

\author{Xiaohui Lin}
\orcid{0009-0006-1639-5862}
\affiliation{%
  \institution{\normalsize Shandong University}
  \city{Jinan}
  \country{China}
}
\email{202415286@mail.sdu.edu.cn}

\author{Weili Guan}
\orcid{0000-0002-5658-5509}
\affiliation{%
  \institution{\normalsize Harbin Institute of Technology, Shenzhen}
  \city{Shenzhen}
  \country{China}
}
\email{honeyguan@gmail.com}

\renewcommand{\shortauthors}{Tongtong Wang et al.}

\begin{abstract}
InfRared Small Target Detection (IRSTD) is a challenging task. Relying solely on pixel-level information, vision-only methods struggle to distinguish targets from clutter. Current multimodal methods typically describe both targets and backgrounds with a single textual prompt. Such an approach lacks dedicated regional guidance and ignores infrared semantic asymmetry. Consequently, it provides insufficient background suppression information and introduces severe feature optimization conflicts, overwhelming small targets with noise. To address these issues, we propose a novel Asymmetric Dual-text Guided Network (ADGNet). Specifically, accounting for the infrared semantic asymmetry, we first design the Asymmetric Dual-text Prompt (ADP), comprising an image-agnostic abstract target prompt and an image-specific detailed background prompt. To leverage these prompts, we introduce an Asymmetric Dual-Branch Interaction (ADBI) module to separately guide visual features with their respective text priors, protecting targets from noise while suppressing background clutter. Subsequently, we introduce an Adaptive Feature Aggregation (AFA) module to dynamically fuse features from the two branches. Furthermore, we construct a multimodal Asymmetric Image-Text Infrared (AITIR) dataset by providing asymmetric text annotations for three public datasets (IRSTD-1K, NUDT-SIRST, and SIRST). Extensive experiments demonstrate that ADGNet outperforms 21 state-of-the-art (SOTA) methods. Code is available at \url{https://github.com/iLearn-Lab/MM26-ADGNet}.

\end{abstract}





\maketitle

\begin{figure}[t]
  \centering
  \includegraphics[width=0.95\linewidth]{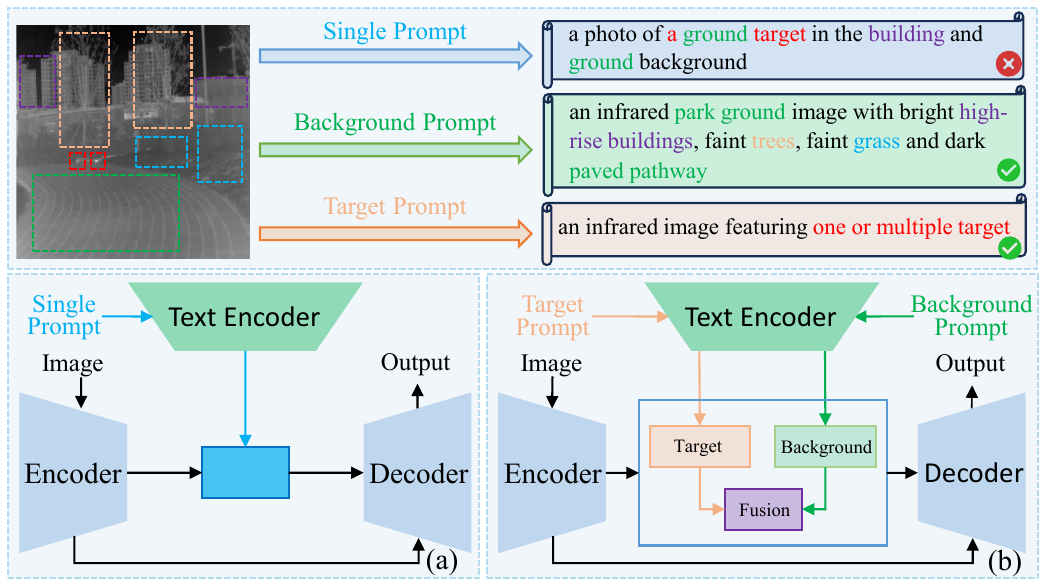}
  \vspace{-10pt}
  \caption{Comparison of text-guided paradigms. (a) Existing method using a single textual prompt. (b) Our proposed ADGNet using two asymmetric textual prompts.}
  \label{fig:intro}
  \vspace{-15pt}
  \Description{The figure compares a conventional single-prompt paradigm with the proposed asymmetric dual-prompt paradigm for infrared small target detection. The upper part shows an infrared image and three textual descriptions: a single prompt that mixes target and background information and is marked as unsuitable, a detailed background prompt, and an abstract target prompt, both marked as suitable. The lower part contains two network diagrams. In (a), a single prompt is encoded and injected into one shared feature block between the image encoder and decoder. In (b), separate target and background prompts guide distinct target and background branches, whose features are fused before being passed to the decoder to generate the output.}
\end{figure}

\section{Introduction}
\label{sec:intro}
InfRared Small Target Detection (IRSTD) aims to accurately segment weak targets from complex backgrounds~\cite{hdnet, fgarnet}. Benefiting from thermal radiation imaging, infrared technology possesses strong anti-interference capabilities in visually obstructed environments. Consequently, it plays an irreplaceable role in maritime monitoring, early warning systems, and autonomous navigation~\cite{mopkl-aaai, mopkl-tgrs, xmz4, surveillance}. However, infrared images exhibit an inherent semantic asymmetry between targets and backgrounds. Infrared targets typically possess singular semantics and severely lack texture or structural priors. In contrast, background regions are rich in clutter and highly structured. This extreme semantic imbalance makes it highly challenging to extract weak thermal signals from clutter, causing severe false alarms and missed detections.

In the field of IRSTD, although early traditional methods (such as filtering~\cite{filter_1, filter_2, filter_3, filter_5,top-hat}, local contrast~\cite{contrast_1, contrast_2, contrast_3, contrast_4}, and low-rank representation~\cite{low_rank_1, low_rank_2, low_rank_3, low_rank_4, non_convex_rank}) provided initial solutions, they relied heavily on hand-crafted features, suffering from limited semantic expression and poor robustness. Subsequently, deep learning methods~\cite{mshnet, dnanet, isnet, uiunet, alcnet, acmnet, irsam, gcinet, pbt, rkformer, abc, dconet, drtenet, fscfnet} have improved detection accuracy by learning deep semantic features end-to-end. However, constrained by a single modality and relying solely on pixel-level information, existing vision-only networks struggle to extract highly discriminative features. Confronted with the semantic asymmetry of infrared images, they still fail to effectively separate textureless targets from complex backgrounds.

In recent years, Vision-Language Models (VLMs)~\cite{glip, detclip, groundingdino}, represented by CLIP~\cite{clip}, have demonstrated cross-modal representation and semantic interaction capabilities~\cite{xmz3,xmz5,lzx7,cui,lzx1,lzx3}. Inspired by this, existing IRSTD methods~\cite{saist, text-irstd} introduce text modality for high-level semantic guidance and cross-modal alignment~\cite{lzx2,lzx5,lm1,lzx4,lzx6}. However, as shown in Fig. \ref{fig:intro}(a), these methods typically describe both targets and backgrounds with a single textual prompt, completely ignoring the semantic asymmetry of infrared images. This paradigm has critical flaws: forcing detailed descriptions onto simple targets introduces semantic noise that leads to false alarms, while oversimplifying the description of complex backgrounds deprives the network of sufficient clutter suppression priors. Furthermore, merging such distinct semantics into a single textual prompt inevitably triggers severe feature optimization conflicts.

To address these challenges, we propose a novel Asymmetric Dual-text Guided Network (ADGNet), whose architectural paradigm is illustrated in Fig. \ref{fig:intro}(b). First, tailored to the semantic asymmetry of infrared images, we design the Asymmetric Dual-text Prompt (ADP). It employs a fixed abstract target prompt to avoid semantic noise, while utilizing a detailed background prompt to provide sufficient priors for clutter suppression. Second, to resolve the feature optimization conflicts caused by a single textual prompt, we propose the Asymmetric Dual-Branch Interaction (ADBI) module. This module explicitly constructs target localization and background suppression branches, independently guiding visual features with their corresponding text priors. Finally, the Adaptive Feature Aggregation (AFA) module dynamically fuses features from both branches, effectively enhancing the target while suppressing background clutter for accurate segmentation in complex scenarios.

The main contributions of this paper are as follows:
\begin{itemize}
   \item We propose a novel ADGNet. Its core ADP overcomes the inherent semantic asymmetry of infrared images. By employing abstract target and detailed background prompts, it avoids semantic noise while providing suppression priors.
   \item We propose a novel ADBI module, which uses corresponding text priors to independently localize targets and suppress clutter. Furthermore, the AFA module dynamically fuses these features, achieving precise target enhancement and effective clutter suppression.
   \item We construct the multimodal AITIR dataset by augmenting the original images of three mainstream public datasets with asymmetric text annotations. Extensive experiments show that ADGNet consistently outperforms 21 existing state-of-the-art (SOTA) methods in complex scenarios.
\end{itemize}

\section{Related Work}
\label{sec:related}
\subsection{Vision-Only Infrared Small Target Detection}
Traditional Infrared Small Target Detection (IRSTD) methods typically rely on hand-crafted features. Specifically, filter-based methods~\cite{filter_2, filter_3} are fast but sensitive to edges and noise. Local contrast-based methods~\cite{contrast_1, contrast_2, contrast_3, contrast_4} highlight targets well but cause high false alarm rates given similar background clutter. Low-rank methods~\cite{low_rank_1, low_rank_2, low_rank_3, low_rank_4} extract targets effectively via matrix separation but are computationally expensive and struggle with complex backgrounds. Overall, traditional methods lack a deep semantic understanding of images, making it difficult to handle real-world environments with complex macro-scenes and local interference. Conversely, deep learning methods extract hierarchical visual features in a data-driven manner, improving detection accuracy~\cite{xmz1,xmz2,xmz6}. 
For example, HDNet~\cite{hdnet} fuses spatial and frequency features, using frequency energy distributions to suppress background interference. To enhance target-background discrimination, SCTransNet~\cite{sctransnet} constructs a cross-transformer network for full-level semantic associations. PKNet~\cite{pknet} introduces a parallel CNN-KAN framework with cyclic interactive fusion for bidirectional local-global feature refinement. Similarly, PQGNet~\cite{pqgnet} leverages wavelet-enhanced reconstruction for improved edge preservation. Furthermore, MPCNet~\cite{mpcnet} incorporates cross-attention feature fusion to enhance target perception, while SP-KAN~\cite{sp-kan} explores a sparse-sine perception KAN network with pattern complementarity modulation to improve nonlinear feature representation.

However, these vision-only methods share a common limitation. Infrared images feature asymmetric semantics: weak targets lack texture and have limited semantics, whereas complex backgrounds contain heavy clutter and rich semantics. Relying solely on pixel-level information, vision-only networks struggle to extract highly discriminative features. As a result, they often fail to distinguish true targets from morphologically similar background clutter.

\subsection{Text-Guided Infrared Small Target Detection}
In recent years, researchers have started to introduce the text modality into IRSTD to provide explicit semantic guidance~\cite{mopkl-aaai, mopkl-tgrs}. By utilizing vision-language models like CLIP~\cite{clip}, these methods use textual prompts to help networks understand complex scenes. 
For example, SAIST~\cite{saist} combines CLIP and SAM, using scene-aware prompts to guide the network in extracting infrared small targets. 
Text-IRSTD~\cite{text-irstd} introduces fuzzy semantic textual prompts to help describe targets and scenes, achieving a preliminary fusion of text and image features via a cross-modal decoder. However, existing text-vision fusion methods usually describe the target and the background together. A detailed description of the target introduces noise and causes false alarms. Meanwhile, a simple description of the background cannot handle complex regions, causing the network to lack sufficient background suppression priors.

To address these limitations, we propose the Asymmetric Dual-text Guided Network (ADGNet). Unlike existing methods that mix target and background semantics within a single prompt, ADGNet leverages an Asymmetric Dual-text Prompt (ADP) to explicitly separate them. Specifically, we employ abstract target prompts to prevent semantic noise and detailed background prompts to extract sufficient clutter suppression priors. To effectively integrate these distinct modalities, we introduce the Asymmetric Dual-Branch Interaction (ADBI) module for independent text-vision guidance, followed by the Adaptive Feature Aggregation (AFA) module to dynamically fuse the interacted representations. This asymmetric paradigm ensures precise target segmentation while maintaining strong robustness against complex clutter.

\begin{figure*}[t]
  \centering
  \includegraphics[width=1.0\textwidth]{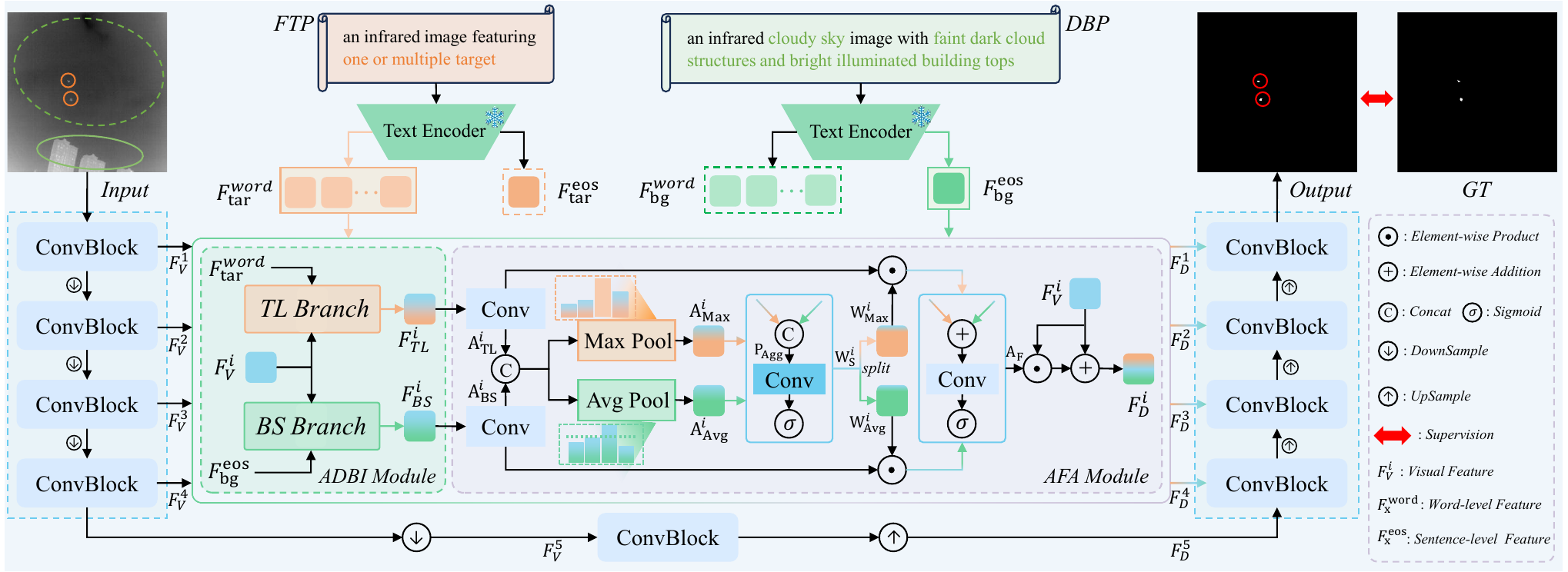}
  \vspace{-20pt}
  \caption{Overall architecture of ADGNet. Within a hierarchical encoder-decoder framework, the ADBI and AFA modules are deployed at the skip connections. By leveraging semantic priors from the FTP and DBP, the ADBI module routes visual features into independent TL and BS branches for asymmetric processing. The AFA module then dynamically aggregates these separated features, effectively resolving semantic conflicts to output a high-precision segmentation mask.}
  \label{fig:adgnet}
  \vspace{-5pt}
  \Description{Overall architecture of ADGNet. Within a hierarchical encoder-decoder framework, the ADBI and AFA modules are deployed at the skip connections. By leveraging semantic priors from the FTP and DBP, the ADBI module routes visual features into independent TL and BS branches for asymmetric processing. The AFA module then dynamically aggregates these separated features, effectively resolving semantic conflicts to output a high-precision segmentation mask.}
\end{figure*}

\begin{figure}[t]
  \centering
  \includegraphics[width=0.90\linewidth]{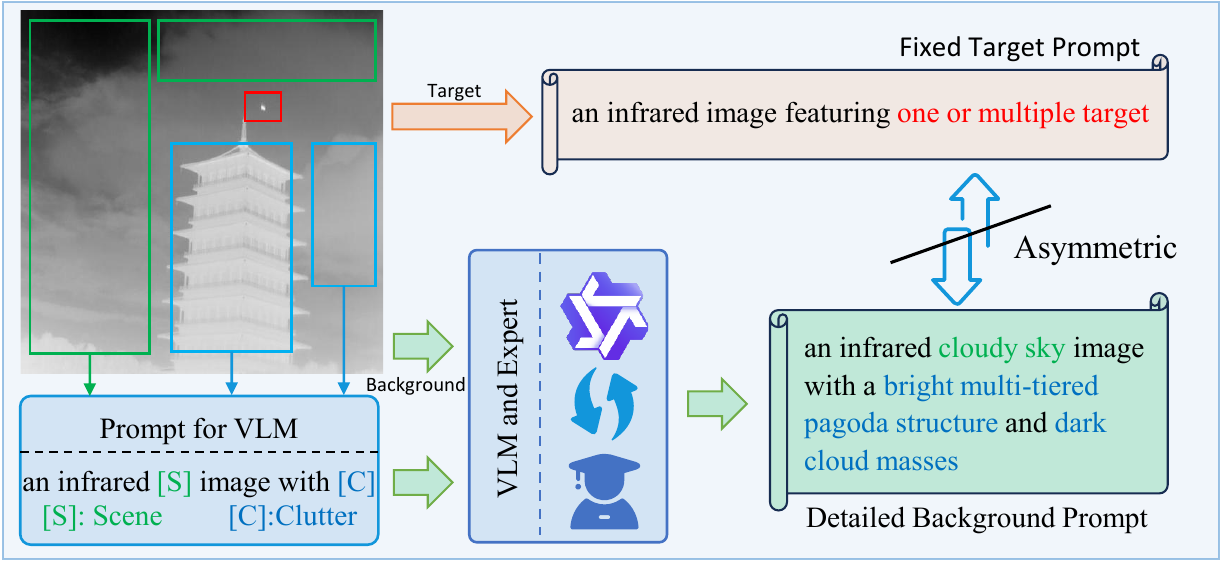}
  \vspace{-5pt}
  \caption{Construction of the Asymmetric Dual-text Prompt.}
  \label{fig:dataset}
  \vspace{-15pt}
  \Description{The figure illustrates the construction of the Asymmetric Dual-text Prompt. An infrared image is decomposed into target and background regions. The target is represented by a fixed, image-agnostic prompt stating that the image contains one or multiple targets. For the background, scene and clutter information are first organized into a structured prompt template and then refined by a vision-language model and expert review to produce an image-specific detailed background description. The resulting fixed target prompt and detailed background prompt form an asymmetric pair for guiding target representation and background modeling.}
\end{figure}

\section{Asymmetric Image-Text Infrared Dataset}
\label{sec:dataset}
Introducing the text modality provides explicit semantic guidance, compensating for the limitations of visual features alone in distinguishing tiny targets from complex clutter. However, existing methods ignore the inherent asymmetric semantics of infrared images and typically use a holistic description for both the target and the background. This approach has obvious flaws: over-describing the tiny target easily introduces semantic noise and causes false alarms, while simply summarizing the complex background leaves the network lacking sufficient priors for clutter suppression.

To address these issues, we propose a novel \textbf{Asymmetric Dual-text Prompt (ADP)}. Based on the asymmetry of infrared images, we design an image-independent abstract target prompt and an image-dependent detailed background prompt to separately guide visual features. Specifically, we use a fixed prompt to streamline the target description, avoiding the introduction of noise. We define this Fixed Target Prompt (FTP) simply as: `an infrared image featuring one or multiple target'. Conversely, to extract sufficient priors for suppressing background clutter, we design a Detailed Background Prompt (DBP): `an infrared [S] image with [C]'. This accurately captures the macro scene [S] and local thermal clutter [C] (e.g., `an infrared forest image with bright tree canopies and dark sky'). 

As shown in Fig. \ref{fig:dataset}, keeping original images and masks unchanged, we use Qwen3-VL-Plus~\cite{qwen3} to generate template-guided background descriptions. Through VLM-expert collaborative refinement, we filter meaningless characters and restrict length to 18 words, finalizing the Detailed Background Prompt. Conversely, targets receive an abstract Fixed Target Prompt. This asymmetric process achieves explicit semantic separation, ultimately extending three existing datasets to construct the multimodal Asymmetric Image-Text Infrared (AITIR) dataset.

\section{Method}
\label{sec:method}
\subsection{Overall Architecture}
The overall architecture of ADGNet is shown in Fig. \ref{fig:adgnet}. It employs a multimodal hierarchical encoder-decoder structure. For an input infrared image, the encoder performs feature extraction and downsampling to capture multiscale visual features $F_V^i$. Relying solely on pixel-level information, vision-only models struggle to extract highly discriminative features to separate similar targets from complex clutter. To address this limitation, we propose the Asymmetric Dual-text Prompt (ADP) based on the semantic asymmetry of infrared images, utilizing two distinct and asymmetric prompts to provide clear semantic guidance. A text encoder initialized with the pre-trained CLIP model extracts word-level features $F_{tar}^{word}$ from the Fixed Target Prompt and sentence-level features $F_{bg}^{eos}$ from the Detailed Background Prompt. To avoid semantic interference between target enhancement and background filtering in a single branch, we deploy the Asymmetric Dual-Branch Interaction (ADBI) module at the first four skip connections to guide visual features using the corresponding text features. To integrate these features and maximize the target-to-clutter ratio, we propose the Adaptive Feature Aggregation (AFA) module for dynamic feature fusion. The fused features are then concatenated with the upsampled representations from the previous decoder layer and passed to the subsequent decoder stage. Finally, the decoder restores spatial resolution and refines the features through progressive upsampling to output the final high-precision target prediction mask $P$.

\subsection{ADBI Module}
Existing multimodal methods typically process the holistic description of the infrared target and background within a single interaction process. However, in complex infrared scenarios, target enhancement and background suppression are mutually restrictive at the feature level. Excessive background filtering easily erases weak targets, while naive target enhancement often simultaneously amplifies similar thermal clutter. To break this feature-level trade-off and perform these two conflicting tasks separately, we propose the Asymmetric Dual-Branch Interaction (ADBI) module.

\begin{figure}[t]
  \centering
  \includegraphics[width=0.90\linewidth]{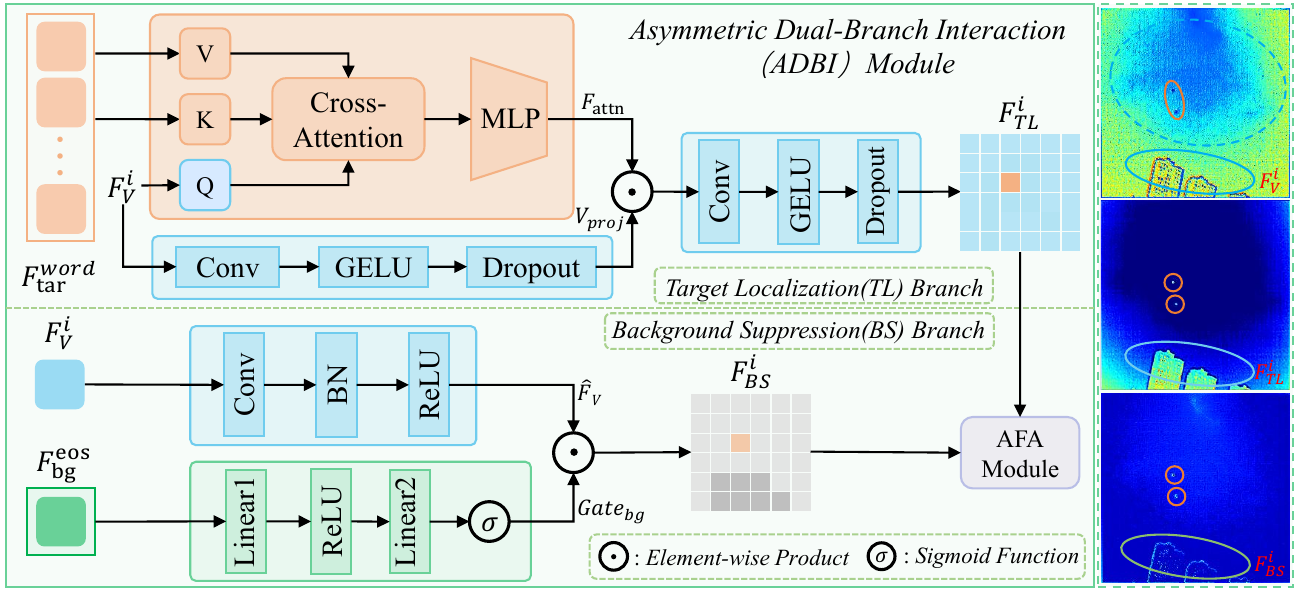}
  \vspace{-10pt}
  \caption{Detailed structure of the ADBI module. The TL and BS branches leverage word-level priors and sentence-level semantics to highlight targets and filter clutter. The right heatmaps visualize the original and refined features.}
  \label{fig:adbi}
  \vspace{-15pt}
  \Description{The figure illustrates the Asymmetric Dual-Branch Interaction (ADBI) module, which contains a Target Localization (TL) branch and a Background Suppression (BS) branch. In the TL branch, the visual feature serves as the query, while word-level target features provide the key and value for cross-attention. The resulting attention feature is processed by an MLP, multiplied with a projected visual feature, and refined by convolution, GELU, and dropout to produce the target-localized feature. In the BS branch, the visual feature is refined by convolution, batch normalization, and ReLU, while the sentence-level background feature is transformed by two linear layers and a sigmoid function to generate a background gate. Element-wise multiplication suppresses background clutter and produces the background-suppressed feature. The outputs of the two branches are then passed to the AFA module. Heatmaps on the right visualize the original visual feature, the target-localized feature, and the background-suppressed feature.}
\end{figure}

As shown in Fig. \ref{fig:adbi}, the ADBI module designs two independent branches to separately guide visual features using specific text priors. Specifically, the Target Localization (TL) branch uses word-level features $F_{tar}^{word} \in \mathbb{R}^{512 \times 20}$ from the FTP to locate the target. This ensures that weak targets are protected from noise interference. Meanwhile, the Background Suppression (BS) branch uses sentence-level features $F_{bg}^{eos} \in \mathbb{R}^{512}$ from the DBP to suppress complex background clutter. By guiding visual features with specific text priors in independent pathways, this asymmetric design successfully avoids interference between target enhancement and background filtering. 

\textbf{TL branch.} This branch aims to precisely locate weak targets. Taking the first-layer visual feature $F_V^1 \in \mathbb{R}^{C \times H \times W}$ (where $C=16, H=256, W=256$) as an example, it is first spatially flattened into $F_{flatten} \in \mathbb{R}^{C \times HW}$, and subsequently projected to obtain the local visual representation $V_{proj} \in \mathbb{R}^{C \times HW}$, as formulated in Eq. \ref{eq:vproj}:
\begin{equation}
  \label{eq:vproj}
  V_{proj} = \text{Dropout}(\text{GELU}(\text{Conv}(F_{flatten}))),
\end{equation}
where $\text{Dropout}(\cdot)$ prevents overfitting, and $\text{GELU}(\cdot)$ denotes the Gaussian Error Linear Unit. Unlike ReLU, GELU provides a smooth nonlinearity. This preserves near-zero feature variations, preventing the loss of faint thermal signatures during projection.

Infrared small targets are extremely tiny and sparse. Conventional global operations aggregate spatial dimensions, which dilutes the target signal and destroys its precise pixel-level coordinates. To overcome this, we employ a Cross-Attention mechanism to perform dense, pixel-level interactions between visual features and target semantics. This ensures accurate localization while strictly preserving high-resolution spatial details. First, the flattened feature $F_{flatten}$ and the word-level feature $F_{tar}^{word}$ are mapped to obtain the query $Q$, key $K$, and value $V$, as formulated in Eq. \ref{eq:q} and Eq. \ref{eq:kv}:
\begin{equation}
  \label{eq:q}
  Q = \text{InstanceNorm}(\text{Conv}_Q(F_{flatten})), 
\end{equation}
\begin{equation}
  \label{eq:kv}
  K = \text{Conv}_K(F_{tar}^{word}), V = \text{Conv}_V(F_{tar}^{word}), 
\end{equation}
where $\text{Conv}_Q$, $\text{Conv}_K$, $\text{Conv}_V$ are distinct convolutional operations. $\text{InstanceNorm}(\cdot)$ normalizes spatial contrast to resist absolute intensity variations. This generates the visual query $Q \in \mathbb{R}^{C \times HW}$, and the text-derived key and value $K, V \in \mathbb{R}^{C \times 20}$.

Next, we compute the cross-modal similarity map and apply a convolutional projection to obtain the target-enhanced feature $F_{attn}\in \mathbb{R}^{C \times HW}$, as formulated in Eq. \ref{eq:lang}:
\begin{equation}
\label{eq:lang}
F_{attn} = Proj([(\text{Softmax}\left(\frac{Q^TK}{\sqrt{d_k}}\right)V^T)]^{T}), 
\end{equation}
where $d_k$ is the scaling factor, $Proj(\cdot) = \text{InstanceNorm}(Conv(\cdot))$. 

Using the target prior, this mechanism computes pixel-level spatial responses to highlight the target without introducing noise. Finally, we modulate the visual feature $V_{proj}$ with $F_{attn}$ to obtain the target localization feature $F_{TL}\in \mathbb{R}^{C \times HW}$, as formulated in Eq.~\ref{eq:FTL}:
\begin{equation}
\label{eq:FTL}
F_{TL} = \text{Dropout}(\text{GELU}(\text{Conv}(V_{proj} \odot F_{attn}))), 
\end{equation}
where $\odot$ denotes element-wise product, and the final $F_{TL}$ is reshaped to $\mathbb{R}^{C \times H \times W}$. As shown in the feature heatmaps on the right side of Fig. \ref{fig:adbi}, this operation efficiently activates target pixels, preventing weak targets from being erased and ensuring that the localization process remains free from noise interference.

\textbf{BS branch.} This branch utilizes global semantic information to suppress complex clutter. The visual feature $F_V^1 \in \mathbb{R}^{C \times H \times W}$ is processed by a convolutional block to obtain the projected feature $\hat{F}_V  \in \mathbb{R}^{C \times H \times W}$, as formulated in Eq. \ref{eq:FV1}:
\begin{equation}
    \hat{F}_V = \text{ReLU}(\text{BN}(\text{Conv}(F_V^1))), 
    \label{eq:FV1}
\end{equation}
where BN is Batch Normalization and ReLU is activation function.

Since background clutter exhibits a global structure, we utilize a global channel-wise gating mechanism. The sentence-level feature $F_{bg}^{eos}$ captures the macro semantics of the scene. It is fed into a Text Gate to generate the modulation weight, as formulated in Eq. \ref{eq:gate}:
\begin{equation}
  \label{eq:gate}
  Gate_{bg} = \text{Sigmoid}(\text{Linear}_2(\text{ReLU}(\text{Linear}_1(F_{bg}^{eos})))), 
\end{equation}
where $\text{Linear}_1$ and $\text{Linear}_2$ are fully connected layers. The resulting $C$-dimensional vector is reshaped into a tensor in $\mathbb{R}^{C \times 1 \times 1}$ to form $Gate_{bg}$, which is spatially broadcast to match the visual feature.

Projecting background prompt semantics into the channel dimension adaptively recalibrates feature maps by suppressing channels activated by thermal clutter while preserving spatial context. We then modulate $\hat{F}_V$ with $Gate_{bg}$ to obtain the background-suppressed feature $F_{BS}\in \mathbb{R}^{C \times H \times W}$, as in Eq.~\ref{eq:BS}:
\begin{equation}
  \label{eq:BS}
  F_{BS} = \hat{F}_V \odot Gate_{bg}, 
\end{equation}
where $\odot$ denotes element-wise product. As shown in the feature heatmaps on the right side of Fig. \ref{fig:adbi}, this operation effectively suppresses complex clutter responses, yielding a clean feature representation that significantly reduces false alarms.

In summary, the ADBI module separately guides visual features via respective text priors to suppress global clutter and localize targets, achieving effective target-background separation.

\begin{table*}[!t]
	\renewcommand{\arraystretch}{0.82}
	\setlength{\tabcolsep}{6pt}
	\caption{Quantitative comparisons of our ADGNet and 21 SOTA methods on the AITIR dataset (IRSTD-1K, NUDT-SIRST, and SIRST images with asymmetric text annotations) in terms of IoU(\%), ${\rm P_d}$(\%) and ${\rm F_a}$($10^{-6}$). Best results are in bold.}
	\vspace{-10pt}
	\resizebox{1.0\linewidth}{!}{
		\begin{tabular}{lcc ccc ccc ccc}
			\hline
			\multicolumn{1}{c}{\multirow{2}{*}[-0.7ex]{Method}} & \multicolumn{1}{c}{\multirow{2}{*}[-0.7ex]{Publication}} & \multicolumn{1}{c}{\multirow{2}{*}[-0.7ex]{Modality}} & \multicolumn{3}{c}{\raisebox{-0.4ex}{IRSTD-1K}} & \multicolumn{3}{c}{\raisebox{-0.4ex}{NUDT-SIRST}} & \multicolumn{3}{c}{\raisebox{-0.4ex}{SIRST}}\\
			\cmidrule(lr){4-6} \cmidrule(lr){7-9} \cmidrule(lr){10-12}
			& & & IoU$\uparrow$ & ${\rm P_d}\uparrow$ & ${\rm F_a}\downarrow$ & IoU$\uparrow$ & ${\rm P_d}\uparrow$ & ${\rm F_a}\downarrow$& IoU$\uparrow$ & ${\rm P_d}\uparrow$ & ${\rm F_a}\downarrow$\\
			\hline 
			RIPT \cite{low_rank_1} & JSTARS'17 & Image & 14.11 & 77.55 & 28.31 & 29.44 & 91.85 & 344.3 & 16.79 & 69.76 & 59.33\\
			NRAM \cite{non_convex_rank} & RS'18 & Image & 15.25 & 70.68 & 16.93 & 6.93 & 56.40 & 19.27 & 12.16 & 74.52 & 13.85\\
			PSTNN \cite{low_rank_4} & RS'19 & Image & 24.57 & 71.99 & 35.26 & 14.85 & 66.13 & 44.17 & 30.30 & 72.80 & 48.99\\
			WSLCM \cite{contrast_2} & GRSL'20 & Image & 3.45 & 72.44 & 6619 & 2.28 & 56.82 & 1309 & 6.39 & 88.74 & 4462\\
			\hdashline
			MDvsFA \cite{mdvsfa} & ICCV'19 & Image & 37.34 & 83.71 & 88.52 & 35.86 & 85.22 & 95.37 & 60.30 & 89.35 & 56.35 \\
			ALCNet \cite{alcnet} & TGRS'21 & Image & 65.68 & 89.25 & 27.71 & 72.89 & 96.19 & 30.40 & 73.74 & 97.25 & 26.79 \\
			ISNet \cite{isnet} & CVPR'22 & Image & 61.85 & 90.24 & 31.56 & 81.24 & 97.78 & 6.34 & 70.49 & 95.06 & 67.98 \\
			DNANet \cite{dnanet} & TIP'22 & Image & 65.71 & 91.84 & 17.61 & 88.19 & 98.62 & 9.00 & 77.76 & 96.33 & 10.29 \\
			UIU-Net \cite{uiunet} & TIP'23 & Image & {68.69} & 91.25 & 13.48 & 75.91 & 96.83 & 18.61 & 77.53 & 92.40 & 9.33 \\
			RDIAN \cite{rdian} & TGRS'23 & Image & 59.94 & 87.21 & 33.31 & 82.42 & 96.72 & 14.85 & 70.74 & 95.06 & 48.16 \\
			MTU-Net \cite{mtunet} & TGRS'23 & Image & 64.09 & 90.48 & 12.15 & 77.98 & 96.08 & 17.51 & 74.85 & {99.08} & {7.09} \\
			MSHNet \cite{mshnet} & CVPR'24 & Image & 67.68 & 92.89 & 12.69 & 80.55 & 97.99 & 11.77 & 73.50 & 97.25 & 31.05 \\
			${\rm L^2SKNet}$ \cite{l2sknet} & TGRS'25 & Image & 67.81 & 90.24 & 17.46 & {93.58} & 97.57 & 5.33 & 73.43 & 98.17 & 20.82\\
			MMLNet \cite{mmlnet} & TGRS'25 & Image & 67.75 & 90.82 & 14.68 & 86.79 & 98.52 & 12.78 & {78.94} & 95.41 & 8.16 \\
			BGM \cite{bgm} & TGRS'25 & Image & {69.23} & 91.50 & 11.39 & 93.33 & 98.84 & 5.86 & 76.17 & 98.17 & 12.42 \\
			DRPCA-Net \cite{drpcanet} & TGRS'25 & Image & 66.33 & 91.07 & 16.93 & {93.33} & 99.15 & 6.05 & 72.82 & 98.77 & 9.23 \\
			PKNet \cite{pknet} & TGRS'25 & Image & 68.28 & 91.16 & 10.17 & {94.26} & 97.99 & 3.15 & 79.10 & 98.94 & 16.68 \\
			PQGNet \cite{pqgnet} & TGRS'26 & Image & 69.88 & 92.78 & 6.68 & {93.67} & 98.41 & 7.35 & 80.61 & 99.08 & 13.72 \\
			MPCNet \cite{mpcnet} & TGRS'26 & Image & 68.10 & 91.76 & 7.97 & {92.90} & 98.84 & 5.19 & 80.10 & 99.08 & 5.50 \\
			SP-KAN \cite{sp-kan} & ISPRS'26 & Image & 69.36 & 92.52 & 12.30 & {93.33} & 98.10 & 8.25 & 78.19 & 97.25 & 9.76 \\
			\hdashline
			SAIST \cite{saist} & CVPR'25 & Image+Text & {72.14} & \textbf{96.18} & 4.76 & 95.23 & 99.28 & \textbf{1.31} & 80.82 & 99.56 & \textbf{0.87} \\
			\cellcolor[HTML]{E8F5E9}\textbf{ADGNet (Ours)} & \cellcolor[HTML]{E8F5E9} & \cellcolor[HTML]{E8F5E9}Image+Text & \cellcolor[HTML]{E8F5E9}\textbf{72.38} & \cellcolor[HTML]{E8F5E9}{93.20} & \cellcolor[HTML]{E8F5E9}\textbf{4.10} & \cellcolor[HTML]{E8F5E9}\textbf{95.53} & \cellcolor[HTML]{E8F5E9}\textbf{99.47} & \cellcolor[HTML]{E8F5E9}{2.64} & \cellcolor[HTML]{E8F5E9}\textbf{83.08} & \cellcolor[HTML]{E8F5E9}\textbf{100} & \cellcolor[HTML]{E8F5E9}{4.97} \\
			\hline
		\end{tabular}
	}
	\vspace{-12pt}
	\label{tab:main_quantitative_result}
\end{table*}

\subsection{AFA Module}
As shown in Fig. \ref{fig:adgnet}, the AFA module evaluates spatial importance to dynamically fuse the two text-guided features, thereby simultaneously enhancing the target and suppressing background clutter. Convolutional operations are applied to $F_{TL}$ and $F_{BS}$ to align channel dimensions, yielding $A_{TL}$ and $A_{BS}$, as formulated in Eq. \ref{eq:ABSTL}:
\begin{equation}
  \label{eq:ABSTL}
  A_{TL} = \text{Conv}(F_{TL}), A_{BS} = \text{Conv}(F_{BS}).
\end{equation}

Next, $A_{TL}\in \mathbb{R}^{(C/2) \times H \times W}$ and $A_{BS}\in \mathbb{R}^{(C/2) \times H \times W}$ are concatenated along the channel dimension. We separately apply max pooling and average pooling to the concatenated feature to extract spatial statistics. A $7 \times 7$ convolution processes the aggregated feature $P_{Agg}\in \mathbb{R}^{2 \times H \times W}$ to generate a two-channel dynamic spatial attention weight $W_{S} \in \mathbb{R}^{2 \times H \times W}$, as formulated in Eq. \ref{eq:pagg} and Eq. \ref{eq:Wspatial}:
\begin{equation}
  \label{eq:pagg}
  P_{Agg} = [\text{MaxPool}([A_{TL}; A_{BS}]); \text{AvgPool}([A_{TL}; A_{BS}])], 
\end{equation}
\begin{equation}
  \label{eq:Wspatial}
  W_{S} = \sigma(\text{Conv}_{7 \times 7}(P_{Agg})), 
\end{equation}
where $[\cdot;\cdot]$ denotes channel concatenation, and $\sigma$ is the Sigmoid function. The large receptive field of the $7 \times 7$ convolution helps the network capture broader spatial context to accurately distinguish targets from clutter. 

Subsequently, $W_{S}$ is split along the channel dimension into two independent spatial weights, $W_{Max} \in \mathbb{R}^{1 \times H \times W}$ and $W_{Avg} \in \mathbb{R}^{1 \times H \times W}$. These weights broadcast through the channel dimension to adaptively modulate $A_{TL}$ and $A_{BS}$. The fused representation is then processed by a convolutional layer to yield the final attention map $A_{F}$, as formulated in Eq. \ref{eq:final}:
\begin{equation}
  \label{eq:final}
  A_{F} = \sigma(\text{Conv}(A_{TL} \odot W_{Max} + A_{BS} \odot W_{Avg})), 
\end{equation}
where $A_{F}\in \mathbb{R}^{C \times H \times W}$, and $\odot$ denotes element-wise product.

Finally, $A_{F}$ modulates the original visual feature $F_V^1$ in a residual manner, as formulated in Eq. \ref{eq:out}:
\begin{equation}
  \label{eq:out}
  F_{D} = F_V^1 \odot (1 + A_{F}).
\end{equation}

This residual multiplication preserves the original visual representation. By dynamically fusing the two text-guided features, the AFA module enhances the target and suppresses background clutter. Ultimately, this provides highly discriminative features for the decoder, achieving precise target segmentation.

\subsection{Loss Function}
To enhance the model's sensitivity to small targets, we employ the SoftIoU loss~\cite{softiou}. It is defined in Eq. \ref{eq:loss}:
\begin{equation}
  \label{eq:loss}
  \mathcal{L}_{\mathrm{SoftIoU}} = 1 - \mathrm{SoftIoU}(P, Y),
\end{equation}
where $P \in [0, 1]^{H \times W}$ represents the predicted probability map, and $Y \in \{0, 1\}^{H \times W}$ is the ground-truth label mask.

\begin{figure*}[t]
  \centering
  \includegraphics[width=0.90\textwidth]{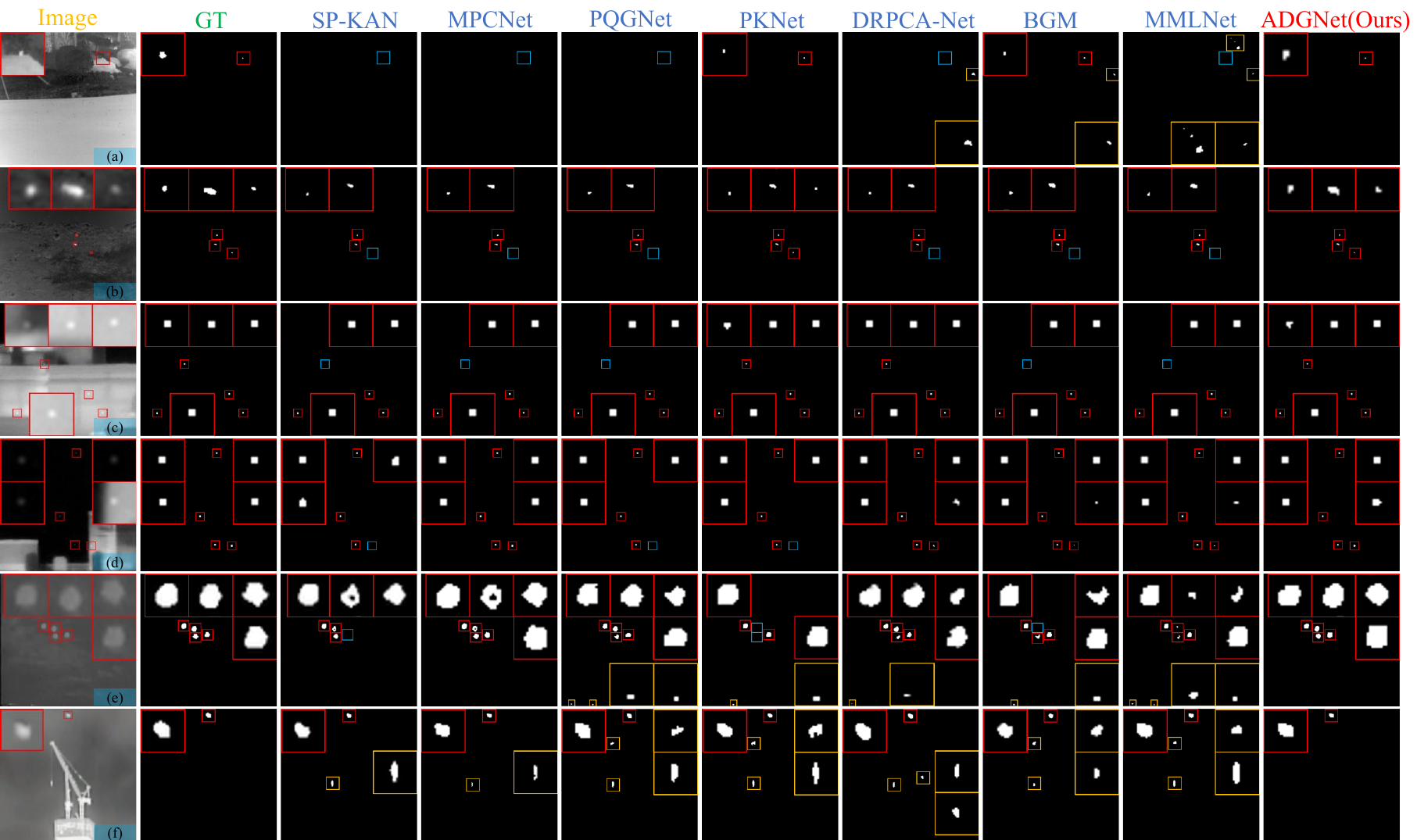}
  \vspace{-10pt}
  \caption{Visual results of different IRSTD methods on the IRSTD-1K (a-b), NUDT-SIRST (c-d), and SIRST (e-f). The boxes in red, yellow, and blue represent correct detections, false alarms, and missed targets, respectively. Enlarged views are in the corners.}
  \label{fig:sota}
  \vspace{-10pt}
  \Description{The figure presents six infrared scenes arranged by dataset: IRSTD-1K in panels (a)-(b), NUDT-SIRST in panels (c)-(d), and SIRST in panels (e)-(f). Each row shows the input image, ground-truth mask, and predictions from SP-KAN, MPCNet, PQGNet, PKNet, DRPCA-Net, BGM, MMLNet, and ADGNet. Red boxes mark correctly detected targets, yellow boxes indicate false alarms caused by background clutter, and blue boxes indicate missed targets. Enlarged insets highlight small or dim target regions. Compared with the other methods, ADGNet produces predictions that more closely match the ground truth, with fewer false alarms and missed targets across the displayed scenes.}
\end{figure*}

\begin{figure}[t]
  \centering
  \includegraphics[width=0.95\linewidth]{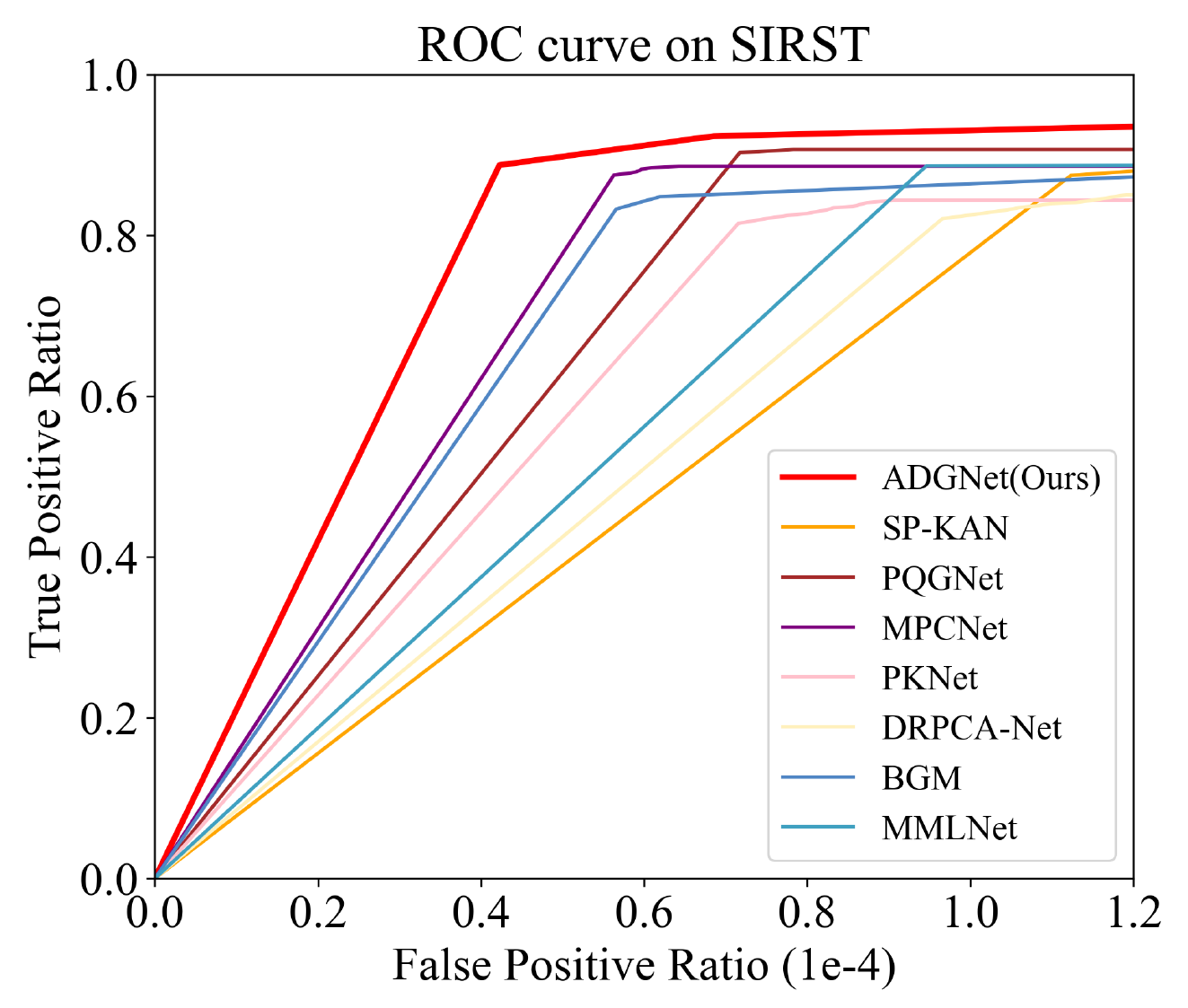}
  \vspace{-5pt}
  \caption{ROC curve on the SIRST dataset.}
  \label{fig:roc_curve}
  \vspace{-15pt}
  \Description{Line graph showing ROC curves for various IRSTD methods on the SIRST dataset. The curve for ADGNet is closest to the top-left corner, indicating the highest True Positive Rate at low False Positive Rates compared to other methods.}
\end{figure}

\section{Experiments}
\label{sec:experiments}
\subsection{Experiment Settings}
\textbf{Datasets and Evaluation Metrics.} 
All experiments are conducted on our newly constructed AITIR dataset, which builds upon three public datasets (IRSTD-1K~\cite{isnet}, NUDT-SIRST~\cite{dnanet}, and SIRST~\cite{acmnet}). Without altering any original images or labels, we exclusively construct asymmetric textual prompts for each image. Following existing works~\cite{hdnet}, IRSTD-1K and SIRST are divided into training and testing sets in the ratio of 4:1, while NUDT-SIRST is equally divided in the ratio of 1:1. To quantitatively evaluate the proposed ADGNet, we employ Intersection over Union ($IoU$) for pixel-level segmentation accuracy, alongside Probability of Detection ($P_d$) and False Alarm Rate ($F_a$) for target-level localization. Furthermore, Receiver Operating Characteristic (ROC) curves are plotted to illustrate the overall dynamic performance.

\textbf{Implementation Details.} 
During experiments, all input images are resized to 256 $\times$ 256, and the pre-trained CLIP-ViT-B/16 is utilized for text encoding. We train ADGNet for 600 epochs using the Adam optimizer with a batch size of 16. The initial learning rate of $5 \times 10^{-4}$ decays by 0.1 at epochs 200 and 400. All implementations are based on PyTorch and run on a single RTX 4090 GPU.

\textbf{Compared methods.} 
We compare our method with 4 traditional methods (RIPT~\cite{low_rank_1}, NRAM~\cite{non_convex_rank}, PSTNN~\cite{low_rank_4}, WSLCM~\cite{contrast_2}), 
16 deep learning methods (MDvsFA~\cite{mdvsfa}, ALCNet~\cite{alcnet}, ISNet~\cite{isnet}, DNANet~\cite{dnanet}, UIU-Net~\cite{uiunet}, RDIAN~\cite{rdian}, MTU-Net~\cite{mtunet}, MSHNet~\cite{mshnet}, ${\rm L^2SKNet}$~\cite{l2sknet}, MMLNet~\cite{mmlnet}, BGM~\cite{bgm}, DRPCA-Net~\cite{drpcanet}, PKNet~\cite{pknet}, PQGNet~\cite{pqgnet}, MPCNet~\cite{mpcnet}, and SP-KAN~\cite{sp-kan}), 
and 1 multimodal method (SAIST \cite{saist}). 
Unlike vision-only methods that only use images, ADGNet utilizes the full image-text dataset. To ensure fairness, all reported results are either cited from original papers or reproduced via official codes under the same settings.

\subsection{Quantitative Comparison}
Table \ref{tab:main_quantitative_result} presents the quantitative comparison of all methods. On the highly challenging IRSTD-1K dataset, ADGNet achieves the highest IoU (72.38\%) and the lowest $F_a$ (4.10). Vision-only methods generally exhibit poor metrics, as relying solely on pixel-level information deprives the network of the semantic priors needed to suppress severe clutter. Furthermore, on the NUDT-SIRST and SIRST datasets, ADGNet attains the highest IoU (95.53\% and 83.08\%) and $P_d$ (99.47\% and 100\%). Although the multimodal method (SAIST) shows competitive $F_a$ on these datasets, it relies exclusively on background descriptions and extracts targets via feature subtraction. Lacking explicit target guidance, this aggressive background suppression easily destroys weak target structures, leading to lower $P_d$ and IoU. In contrast, our ADGNet leverages two asymmetric prompts. Specifically, the detailed background prompt provides sufficient priors to filter clutter, while the simple target prompt offers explicit guidance without introducing the semantic noise that easily causes false detections. Consequently, our model demonstrates superior detection performance and robustness in complex scenarios.

Furthermore, Fig. \ref{fig:roc_curve} illustrates the ROC curves evaluated on the SIRST dataset. As observed, ADGNet reliably secures a higher True Positive Rate (TPR) while strictly bounding the False Positive Rate (FPR). This exceptional TPR-FPR balance further highlights its robust detection capability against current leading methods.

\subsection{Qualitative Comparison}
Fig. \ref{fig:sota} illustrates qualitative comparisons between our ADGNet and seven advanced methods. Existing vision-only models exhibit obvious limitations when facing the extreme semantic asymmetry of infrared images. For instance, in scenes with strong structural interference or bright clutter (e.g., scenes (a), (e) and (f)), these methods lack the necessary semantic priors for background suppression. This deficiency directly leads to severe false alarms. Furthermore, when detecting dim or multiple targets (e.g., scenes (b), (c) and (d)), the absence of explicit target guidance causes them to fail to capture sparse visual features, resulting in frequent missed detections. In contrast, our ADGNet specifically handles this semantic asymmetry. The detailed background prompt provides sufficient priors to effectively filter out complex distractors and suppress severe background clutter. Simultaneously, the simple target prompt offers explicit guidance to accurately localize all weak targets. As shown in the last column, ADGNet consistently achieves fully correct detections and effectively avoids both missed targets and false alarms. This performance demonstrates the robustness and segmentation accuracy of ADGNet in complex scenarios.

\begin{table}[t]
    \centering
    \renewcommand{\arraystretch}{0.95} 
    \setlength{\tabcolsep}{4pt} 
    \caption{Ablation study of the ADBI and AFA modules.}
    \vspace{-8pt}
    \begin{center}
        {\fontsize{7pt}{8pt}\selectfont} 
        \begin{tabular}{lcccccc}
            \hline
            \multicolumn{1}{c}{\multirow{2}{*}[-0.7ex]{Variant}} & \multicolumn{3}{c}{\raisebox{-0.3ex}{IRSTD-1K}} & \multicolumn{3}{c}{\raisebox{-0.3ex}{NUDT-SIRST}} \\
            \cmidrule(lr){2-4} \cmidrule(lr){5-7}
            & IoU $\uparrow$ & ${\rm P_d} \uparrow$ & ${\rm F_a} \downarrow$ & IoU $\uparrow$ & ${\rm P_d} \uparrow$ & ${\rm F_a} \downarrow$ \\
            \hline
            baseline & 64.52 & 86.73 & 22.09 & 86.72 & 93.33 & 18.89 \\
            +ADBI    & 70.82 & 90.14 & 9.49  & 93.13 & 96.19 & 8.94  \\
            +AFA     & 69.43 & 90.48 & 10.25 & 92.86 & 96.51 & 9.56  \\
            \hdashline
            \rowcolor[HTML]{E8F5E9} \textbf{ADGNet (Ours)} & \textbf{72.38} & \textbf{93.20} & \textbf{4.10} & \textbf{95.53} & \textbf{99.47} & \textbf{2.64} \\
            \hline
        \end{tabular}
    \label{tab:Ablation_mkj}
    \vspace{-10pt}
    \end{center}
\end{table}

\begin{figure}[t]
  \centering
  \includegraphics[width=0.95\linewidth]{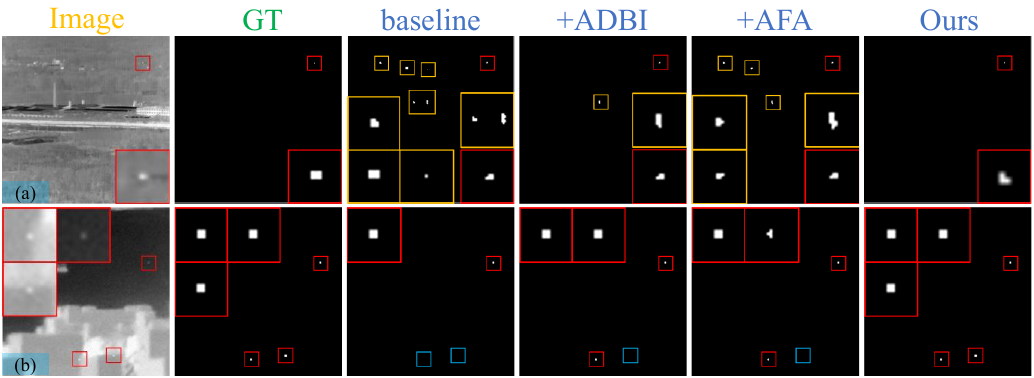}
  \vspace{-5pt}
  \caption{Visual examples of different ADGNet variants.}
  \label{fig:Ablation_mkj}
  \vspace{-12pt}
  \Description{The figure compares two infrared scenes across the input image, ground-truth mask, baseline model, baseline with ADBI, baseline with AFA, and complete ADGNet. The vision-only baseline produces multiple false alarms in the cluttered scene and misses the dim target in the second scene. Adding either ADBI or AFA reduces these errors but leaves residual clutter or incomplete target regions. The complete ADGNet produces predictions that most closely match the ground truth in both scenes.}
\end{figure}

\subsection{Ablation Study}
\textbf{Effectiveness of Proposed Modules. }
To assess the individual contributions of the ADBI and AFA modules, we use a standard vision-only encoder-decoder as the baseline. As shown in Table \ref{tab:Ablation_mkj}, the baseline model is highly vulnerable to background interference without semantic priors. Specifically, on the IRSTD-1K dataset, it produces a high $F_a$ of 22.09 and a low IoU of 64.52\%. Adding the ADBI module alleviates this problem, where IoU increases by 6.30\% and $F_a$ decreases by 12.60. This confirms that the ADBI module can effectively use asymmetric textual prompts to isolate targets from heavy clutter. Similarly, adding the AFA module also brings significant gains. When both modules work together, the full ADGNet achieves the best results. Compared to the baseline on the IRSTD-1K dataset, IoU increases by 7.86\% and $F_a$ decreases by 17.99. Additionally, $P_d$ increases to 99.47\% on the NUDT-SIRST dataset. These quantitative improvements are visually confirmed by our qualitative results. As shown in Fig. \ref{fig:Ablation_mkj}, the vision-only baseline generates many false alarms in scene (a) and completely misses the dim target in scene (b). While adding the ADBI or AFA module alone reduces these errors, some artifacts and incomplete target shapes remain. In contrast, the full ADGNet eliminates false alarms and segments intact targets. This validates that the semantic guidance from the ADBI module and the feature fusion provided by the AFA module improve detection accuracy and robustness in complex scenes.

\begin{figure}[t]
  \centering
  \includegraphics[width=0.95\linewidth]{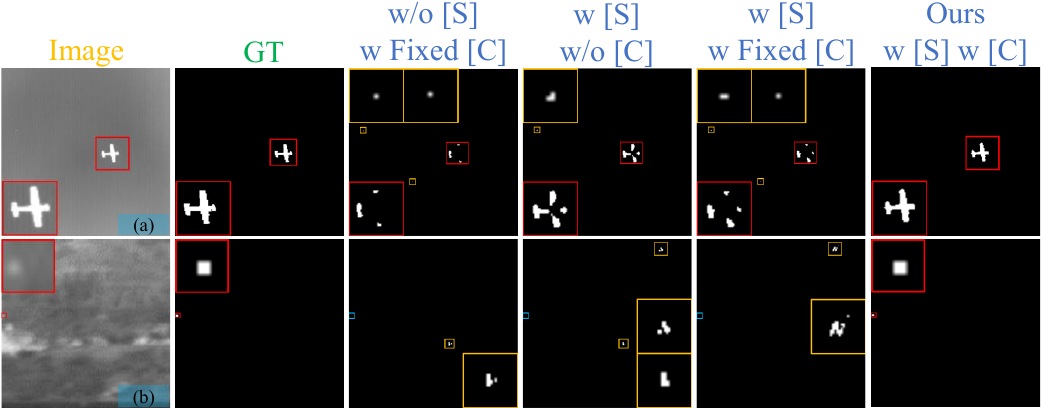}
  \vspace{-5pt}
  \caption{Visual examples of different prompt variants.}
  \label{fig:Ablation_prompt}
  \vspace{-12pt}
  \Description{The figure compares two infrared scenes using four background-prompt variants: a fixed clutter description without scene information, scene information without clutter details, scene information with a fixed clutter description, and the complete image-specific scene-and-clutter prompt. Incomplete prompts produce fragmented targets, missed detections, or false alarms in cluttered regions. The complete prompt preserves the target regions while more effectively suppressing background interference.}
\end{figure}

\textbf{Analysis of Detailed Background Prompt. } 
To investigate the compositional impact of the DBP, we tested different prompt variants based on the standard template: `an infrared [S] image with [C]'. Here, [S] represents the specific macro scene, and [C] denotes the specific local clutter. We defined a general clutter description, Fixed [C]: `complex thermal background clutter and high-contrast distractors'. As shown in Table \ref{tab:Ablation_prompt}, the variant lacking specific scene context (`w/o [S], w Fixed [C]') struggles to handle diverse backgrounds, yielding the lowest IoU of 69.93\% on the IRSTD-1K dataset. Introducing the scene description (`w [S]') improves performance, but missing specific clutter details or using a generic description (`w/o [C]' or `w Fixed [C]') still fails to provide sufficient priors for filtering specific local noise. Conversely, our complete template (`w [S], w [C]') achieves the highest IoU (72.38\% and 95.53\%) and the lowest $F_a$ (4.10 and 2.64) across both datasets. As shown in Fig. \ref{fig:Ablation_prompt}, without precise background priors, the incomplete variants fail to effectively suppress clutter, which interferes with target feature extraction. As a result, the network severely fragments the targets in scene (a). Even worse, in the highly cluttered scene (b), the incomplete variants completely miss the true target and generate severe false alarms. In contrast, our complete template perfectly segments intact targets while entirely eliminating background noise. This confirms that our detailed, image-specific scene and clutter descriptions significantly enhance the robustness of the model.

\begin{table}[t]
    \centering
    \renewcommand{\arraystretch}{0.95} 
    \setlength{\tabcolsep}{3.5pt} 
    \caption{Ablation study of different background prompts.}
    \vspace{-8pt}
    \begin{center}
        {\fontsize{7pt}{8pt}\selectfont} 
        \begin{tabular}{lcccccc}
            \hline
            \multicolumn{1}{c}{\multirow{2}{*}[-0.7ex]{Variant}} & \multicolumn{3}{c}{\raisebox{-0.3ex}{IRSTD-1K}} & \multicolumn{3}{c}{\raisebox{-0.3ex}{NUDT-SIRST}} \\
            \cmidrule(lr){2-4} \cmidrule(lr){5-7}
            & IoU$\uparrow$ & ${\rm P_d}\uparrow$ & ${\rm F_a}\downarrow$ & IoU$\uparrow$ & ${\rm P_d}\uparrow$ & ${\rm F_a}\downarrow$ \\
            \hline
            w/o [S], w Fixed [C] & 69.93 & 90.82 & 9.19 & 90.79 & 95.66 & 9.35 \\
            w [S], w/o [C]       & 70.37 & 91.16 & 7.44 & 92.42 & 97.35 & 7.63 \\
            w [S], w Fixed [C]   & 71.25 & 92.18 & 6.15 & 93.12 & 98.31 & 5.45 \\
            \hdashline
            \rowcolor[HTML]{E2F0D9} \textbf{w [S], w [C] (Ours)} & \textbf{72.38} & \textbf{93.20} & \textbf{4.10} & \textbf{95.53} & \textbf{99.47} & \textbf{2.64} \\
            \hline
        \end{tabular}
    \label{tab:Ablation_prompt}
    \vspace{-10pt}
    \end{center}
\end{table}

\begin{figure}[t]
  \centering
  \includegraphics[width=1.0\linewidth]{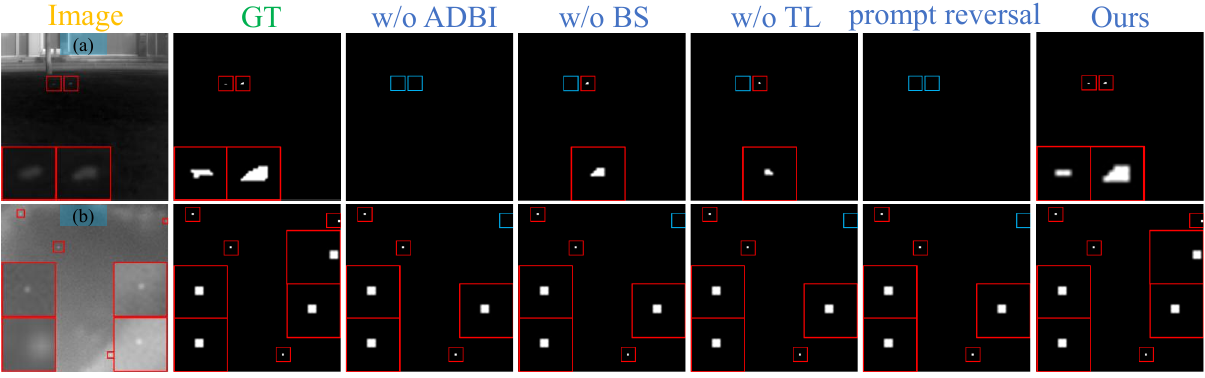}
  \vspace{-15pt}
  \caption{Visual examples of different ADBI module variants.}
  \label{fig:Ablation_adbi}
  \vspace{-10pt}
  \Description{The figure compares two infrared scenes across the input image, ground-truth mask, ADGNet without ADBI, without the Background Suppression branch, without the Target Localization branch, with reversed prompt assignments, and the complete model. Removing the Background Suppression branch or reversing the prompts leaves false responses in cluttered regions, whereas removing the Target Localization branch causes dim targets to be missed. The complete ADBI design produces predictions closest to the ground truth.}
\end{figure}

\textbf{Internal Design of ADBI Module. } 
To verify the internal structure of the ADBI module, we evaluate variants by removing the entire module (`w/o ADBI'), removing individual branches (`w/o BS' and `w/o TL'), and reversing the prompt fusion logic (`Prompt Reversal'). As shown in Table \ref{tab:Ablation_adbi}, completely removing the module severely degrades performance, with IoU decreasing by 2.95\% and $F_a$ increasing by 6.15 on the IRSTD-1K dataset. Removing the BS branch weakens background suppression, where $F_a$ increases by 5.09. Conversely, removing the TL branch harms target extraction, where IoU decreases by 1.42\%. Furthermore, reversing the prompt logic severely confuses the network, causing $F_a$ to increase by 5.47. The visual comparisons in Fig. \ref{fig:Ablation_adbi} further validate these quantitative changes. Specifically, without the BS branch or with reversed prompts, the network fails to filter clutter. Conversely, without the TL branch, the model fails to capture sparse features, completely missing the dim targets. In contrast, the complete ADBI module perfectly segments targets while eliminating noise. This confirms that assigning specific text features to dedicated branches is highly effective for balancing accurate target localization and strict background suppression.

\begin{table}[t]
    \centering
    \renewcommand{\arraystretch}{0.95}
    \setlength{\tabcolsep}{4pt}
    \caption{Ablation study of the ADBI module.}
    \vspace{-8pt}
    \begin{center}
        {\fontsize{7pt}{8pt}\selectfont}
        \begin{tabular}{lcccccc}
            \hline
            \multicolumn{1}{c}{\multirow{2}{*}[-0.7ex]{Variant}} & \multicolumn{3}{c}{\raisebox{-0.3ex}{IRSTD-1K}} & \multicolumn{3}{c}{\raisebox{-0.3ex}{NUDT-SIRST}} \\
            \cmidrule(lr){2-4} \cmidrule(lr){5-7}
            & IoU $\uparrow$ & ${\rm P_d} \uparrow$ & ${\rm F_a} \downarrow$ & IoU $\uparrow$ & ${\rm P_d} \uparrow$ & ${\rm F_a} \downarrow$ \\
            \hline
            w/o ADBI        & 69.43 & 90.48 & 10.25 & 92.86 & 96.51 & 9.56 \\
            w/o BS          & 71.15 & 92.18 & 9.19  & 94.31 & 98.20 & 7.74 \\
            w/o TL          & 70.96 & 91.50 & 6.98  & 93.94 & 97.67 & 6.34 \\
            Prompt Reversal & 70.43 & 91.16 & 9.57  & 93.03 & 97.04 & 8.16 \\
            \hdashline
            \rowcolor[HTML]{E2F0D9} \textbf{ADGNet (Ours)} & \textbf{72.38} & \textbf{93.20} & \textbf{4.10} & \textbf{95.53} & \textbf{99.47} & \textbf{2.64} \\
            \hline
        \end{tabular}
    \label{tab:Ablation_adbi}
    \vspace{-10pt}
    \end{center}
\end{table}

\begin{table}[t]
    \centering
    \renewcommand{\arraystretch}{0.95}
    \setlength{\tabcolsep}{4pt}
    \caption{Ablation study of the AFA module.}
    \vspace{-8pt}
    \begin{center}
        {\fontsize{7pt}{8pt}\selectfont}
        \begin{tabular}{lcccccc}
            \hline
            \multicolumn{1}{c}{\multirow{2}{*}[-0.7ex]{Variant}} & \multicolumn{3}{c}{\raisebox{-0.3ex}{IRSTD-1K}} & \multicolumn{3}{c}{\raisebox{-0.3ex}{NUDT-SIRST}} \\
            \cmidrule(lr){2-4} \cmidrule(lr){5-7}
            & IoU $\uparrow$ & ${\rm P_d} \uparrow$ & ${\rm F_a} \downarrow$ & IoU $\uparrow$ & ${\rm P_d} \uparrow$ & ${\rm F_a} \downarrow$ \\
            \hline
            w/o AFA & 70.82 & 90.14 & 9.49 & 93.13 & 96.19 & 8.94 \\
            w/o Avg & 71.50 & 92.52 & 8.42 & 94.41 & 98.31 & 7.68 \\
            w/o Max & 71.25 & 91.84 & 6.76 & 94.25 & 97.88 & 5.49 \\
            \hdashline
            \rowcolor[HTML]{E2F0D9} \textbf{ADGNet (Ours)} & \textbf{72.38} & \textbf{93.20} & \textbf{4.10} & \textbf{95.53} & \textbf{99.47} & \textbf{2.64} \\
            \hline
        \end{tabular}
    \label{tab:Ablation_afa}
    \vspace{-10pt}
    \end{center}
\end{table}

\begin{figure}[t]
  \centering
  \includegraphics[width=1.0\linewidth]{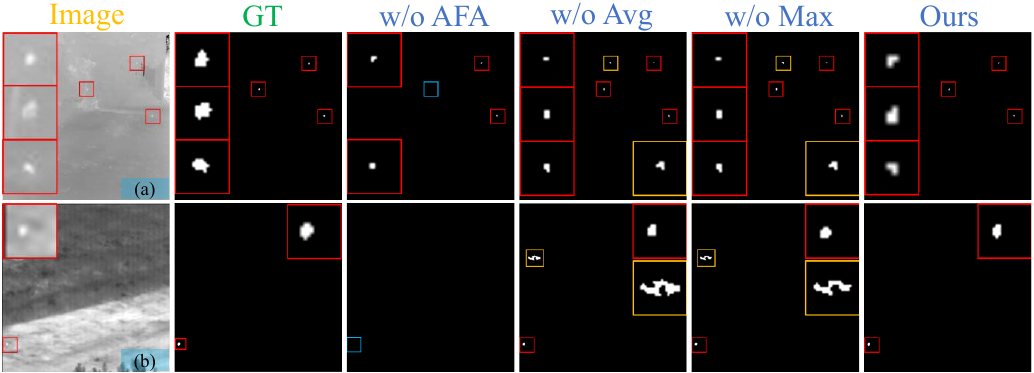}
  \vspace{-15pt}
  \caption{Visual examples of different AFA module variants.}
  \label{fig:Ablation_afa}
  \vspace{-15pt}
  \Description{The figure compares two infrared scenes across the input image, ground-truth mask, ADGNet without AFA, without average pooling, without max pooling, and the complete model. Removing AFA causes false alarms in cluttered regions and missed dim targets, while removing either pooling branch leaves residual artifacts or incomplete target shapes. The complete AFA design yields cleaner and more complete target predictions.}
\end{figure}

\textbf{Internal Design of AFA Module.} 
To explore the fusion mechanism within the AFA module, we evaluate variants by removing the entire module (`w/o AFA'), the average pooling branch (`w/o Avg'), and the max pooling branch (`w/o Max'). As shown in Table \ref{tab:Ablation_afa}, removing the entire module significantly degrades performance, with IoU decreasing by 1.56\% and $F_a$ increasing by 5.39 on the IRSTD-1K dataset. Removing the average pooling branch limits global context modeling, increasing $F_a$ by 4.32. Similarly, removing the max pooling branch discards salient local features, decreasing IoU by 1.13\%. The visual comparisons in Fig. \ref{fig:Ablation_afa} further validate these quantitative changes. Specifically, without the entire module, the network suffers from severe false alarms in cluttered areas and completely misses dim targets. Furthermore, removing individual pooling branches leaves noticeable artifacts and incomplete target shapes. In contrast, the complete AFA module perfectly segments intact targets while eliminating background noise. This confirms that dynamically fusing features through both pooling branches provides robust feature calibration for precise segmentation.

\begin{table}[t]
    \centering
    \renewcommand{\arraystretch}{1.02} 
    \setlength{\tabcolsep}{3.4pt} 
    \caption{Comparison of model complexity between ADGNet and recent SOTA methods from the past two years. `-' denotes that the code is unavailable for evaluation.}
    \vspace{-5pt}
    \begin{center}
        {\fontsize{7pt}{8pt}\selectfont} 
        \begin{tabular}{lcccc}
            \hline
            Method & Year & Params (M) $\downarrow$ & FLOPs (G) $\downarrow$ & FPS (f/s) $\uparrow$ \\
            \hline
            ${\rm L^2SKNet}$~\cite{l2sknet} & 2025 & 0.90 & 6.89 & 85.53 \\
            MMLNet~\cite{mmlnet} & 2025 & 3.58 & 20.41 & 40.44 \\
            BGM~\cite{bgm} & 2025 & 4.08 & 6.77 & 55.04 \\
            DRPCA-Net~\cite{drpcanet} & 2025 & 1.17 & 73.84 & 38.86 \\
            PKNet~\cite{pknet} & 2025 & 9.41 & 15.25 & 44.88 \\
            PQGNet~\cite{pqgnet} & 2026 & 1.19 & 9.89 & 27.30 \\
            MPCNet~\cite{mpcnet} & 2026 & 2.28 & 12.35 & 61.20 \\
            SP-KAN~\cite{sp-kan} & 2026 & 4.40 & 4.71 & 63.84 \\
            \hdashline
            SAIST~\cite{saist} & 2025 & (383.28)+6.29 & - & - \\
            \rowcolor[HTML]{E8F5E9} {ADGNet (Ours)} & & {(37.83)+4.61} & {7.25} & {66.41} \\
            \hline
        \end{tabular}
    \label{tab:complexity}
    \vspace{-15pt}
    \end{center}
\end{table}

\subsection{Computational Efficiency}
We evaluate model efficiency using parameters (Params), floating-point operations (FLOPs), and frames per second (FPS). As shown in Table \ref{tab:complexity}, for multimodal methods, we separately report pre-trained model and core network parameters. First, compared to SAIST (whose Params are directly sourced from its original paper \cite{saist}), ADGNet requires fewer core network parameters (4.61 M vs. 6.29 M) and is substantially more lightweight overall (42.44 M vs. 389.57 M). Second, our core network (4.61 M) remains highly competitive even among vision-only methods. Although incorporating the pre-trained CLIP text encoder (37.83 M) increases our total parameters, the actual computational overhead remains highly manageable. Specifically, ADGNet achieves 66.41 FPS with 7.25 G FLOPs, demonstrating competitive inference efficiency compared with lightweight models such as DRPCA-Net (38.86 FPS) and SP-KAN (63.84 FPS). While $L^2$SKNet (85.53 FPS) exhibits higher throughput, ADGNet provides significantly better detection accuracy, striking an excellent balance between performance and efficiency.

\section{Conclusion}
\label{sec:conclusion}
In this paper, we propose ADGNet to solve the semantic bottlenecks of vision-only methods and the limitations of symmetric text-guided methods. To handle the semantic asymmetry of infrared images, we design an Asymmetric Dual-text Prompt to achieve semantic separation by using fixed target and detailed background descriptions. This prevents noise while providing enough priors for background suppression. To resolve feature optimization conflicts, our ADBI module explicitly assigns the corresponding text priors to independent branches, separately guiding visual features for target localization and background suppression. Next, the AFA module dynamically fuses these features to completely filter clutter and enhance targets. Finally, we build the multimodal AITIR dataset with high-quality asymmetric text annotations. Extensive experiments show that ADGNet effectively eliminates severe false alarms and missed detections, achieving state-of-the-art (SOTA) performance.

\clearpage
\begin{acks}
This work was supported in part by the National Natural Science Foundation of China (NSFC) under Grant 62576194, in part by the ``Key R\&D Program of Shandong Province, China'' under Grant 2025CXGC020101, and in part by the project Youth Science Fund (B) supported by Shandong Provincial Natural Science Foundation under Grant ZR2026QB12.
\end{acks}


\bibliographystyle{ACM-Reference-Format}
\bibliography{ADGNet}


\end{document}